# Real Time Incremental Image Mosaicking Without Use of Any Camera Parameter

Süleyman Melih Portakal
*Wireless DC Department*
*Huawei*
Istanbul, Turkey
suleyman.melih.portakal@huawei.com

Ahmet Alp Kındıroğlu
*Wireless DC Department*
*Huawei*
Istanbul, Turkey
ahmet.alp.kindiroglu@huawei.com

Mahiye Uluyağmur Öztürk
*Wireless DC Department*
*Huawei*
Istanbul, Turkey
mahiye.uluyagmur.ozturk@huawei.com

*Abstract*— Over the past decade, there has been a significant increase in the use of Unmanned Aerial Vehicles (UAVs) to support a wide variety of missions, such as remote surveillance, vehicle tracking, and object detection. For problems involving processing of areas larger than a single image, the mosaicking of UAV imagery is a necessary step. Real-time image mosaicking is used for missions that requires fast response like search and rescue missions. It typically requires information from additional sensors, such as Global Position System (GPS) and Inertial Measurement Unit (IMU), to facilitate direct orientation, or 3D reconstruction approaches to recover the camera poses. This paper proposes a UAV-based system for real-time creation of incremental mosaics which does not require either direct or indirect camera parameters such as orientation information. Inspired by previous approaches, in the mosaicking process, feature extraction from images, matching of similar key points between images, finding homography matrix to warp and align images, and blending images to obtain mosaics better looking, plays important roles in the achievement of the high quality result. Edge detection is used in the blending step as a novel approach. Experimental results show that real-time incremental image mosaicking process can be completed satisfactorily and without need for any additional camera parameters.

*Keywords—UAV, Incremental mosaicking, Real time mosaicking, SIFT, Homography Transformation*

## I. Introduction

In recent years, Unmanned aerial vehicle (UAV)-based vision systems have been increasingly used to assist both military and civilian operations in monitoring vast areas that are challenging for human operators to handle. Some examples of these missions include land monitoring [1], search and rescue (SAR) [2], military operations, and some agricultural operations. Examples of usage of UAVs in the military arena include the routine surveillance of important regions near army bases, connecting roadways, and refugee camps in order to detect the existence of devices that may endanger the passage of army and humanitarian convoys. In the field of agriculture, making appropriate use of UAVs in regions that include plantations enables the performance of activities such as monitoring plant health, counting plants, and conducting assessments after extreme weather events such as floods [3-4].

A method known as aerial picture mosaicking is applicable to a wide range of activities, such as land use planning, agricultural management, forest management, conservation efforts, and urban planning, among others. When using a camera that is mounted to a UAV, just a small portion of the overall scene may be captured in a single photograph. When using digital earth applications, it is often necessary to stitch together hundreds or even thousands of images in order to generate a bigger image that may successfully cover desired areas of interest in a continuous figure.

Using the approach of aerial image mosaicking, a single large image is created from a collection of smaller images acquired over the same location. To perform mosaicking of images belonging to the real world, the most common approach is to map them to a 3D surface model. For satellite mosaicking applications that cover very large areas on earth, taking into account earths shape becomes necessary. However for UAV images, much faster computation is possible if it is assumed that all the obtained images are on the same 2D plane [23]. When performing aerial mapping or surveying tasks, UAVs collect photos of the target region in a systematic matter. Since the area covered by a single image is limited by the cameras field of view, resulting collection of images are combined into a mosaic for applications that require images covering larger areas. Both incremental and non-incremental methods can be utilized while constructing a mosaic. In the first approach, the process of creating the mosaic is carried out frame by frame. Beginning with the initial frame of a flight permits real-time operations while a UAV is flying over a specified area. The second approach, on the other hand, needs that the full image collection be captured before it can be processed into a mosaic. The latter choice inhibits real-time actions.

Real-time image mosaicking can aid tasks that demand both efficiency and precision, such as urban surveillance and fire monitoring. UAV photography generally requires extra data, such as camera calibration settings, location and orientation data from GPS, or a reference map, so that the mosaicking results can be more accurate. In this article, we provide a method for real-time incremental picture mosaicking that does not require any camera calibration settings, camera postures, or data from GPS. Our method is completely independent of these factors. Using only the unprocessed images collected from a UAV, our technology is able to automatically create results that are visually pleasant to the eye.

The primary purpose of this study is to provide a method for real-time incremental picture mosaicking that does not make use of drone settings. SIFT (Scale Invariant Feature Transform) [5] is a computer vision method that can recognize, identify, and extract feature points from pictures. The Brute Force Matcher (BFM) algorithm is utilized in the process of finding a matching. Following the removal of ouliers by using the RAndom SAmple Consensus (RANSAC) [6], the homography transformation matrix is computed.

It is possible for there to be a difference in the brightness and contrast of the overlapping region compared to the remainder of the mosaic. In order to reduce the effect of this problem on mosaicking accuracy and make overlapping regions look smoother, the proposed algorithm adopts a method that is known as alpha blending [15]. Alpha blending, however, can make an image hazy and blurry. In order to avoid distortion in the alpha blending process, the proposed method suggests a novel technique that involves first detecting regions in an overlapped area that have more complex edges, like trees. Then new image's pixel weight is increased in these areas. Therefore, results in the complex areas are less unclear.

## II. RELATED WORKS

In the literature, there are numerous examples of producing a mosaic in offline mode, which occurs when all of the frames are accessible for the processing. There are two examples that have been described in [7] and [8], where the authors provide a reliable system that makes use of a SIFT extractor and homography transformation that is based on RANSAC.

When it comes to the real-time processing, the authors of [10] use ORB as the feature extractor and give a temporal and spatial filter in order to get rid of the vast majority of outlier points. SIFT is used as the feature extractor in [11], and Euclidean distance is used to match the frames. The authors of [12] use an incremental method together with additional UAVs to cover a region of interest and construct a descriptive mosaic.

Real-time incremental georeferenced picture mosaicking is also introduced in [13]. The algorithm in [13] applies a ROI in order to decrease the amount of computing necessary for the stitching of each new frame onto the mosaic. This enables the algorithm to speed up all of the stages that are involved in the process of stitching. It utilizes A-KAZE as a feature descriptor, implements ROI in order to accelerate the stitching process, and uses rigid transformation instead of homography transformation. Our proposed mosaicking method, which was inspired of this work, in the feature extraction step, uses a portion of the mosaic that is 3 times the size of the last stitched image in order to make feature extraction and matching steps faster.

Whether the mosaicking is performed in real time or not, most of the methods involve at least one of the following: camera calibration settings; camera postures; or data from GPS and IMU. A counter example is [14] where there is no utilization of camera pose information in [14]. However, the image mosaicking process in [14] is not real time. Algorithm in it determines the transformation matrix for each picture based on how it corresponds to the others. Unlike [14], proposed method involves working in real time and incremental. Our proposed method is entirely image based. It does not require GPS and IMU data, GCPs, or satellite images as a reference.

## III. PROPOSED METHOD

Fig. 1 is a visualization of the pipeline that the proposed mosaicking method uses. The method is comprised of four primary stages, which are referred to as the feature extraction, matching, finding homography transformation, and blending.

Because the size of the mosaic grows over time, the process of extracting features and matching them typically gets overwhelming after some time has passed. The problem is intended to be avoided by using a method that uses a portion of the mosaic. This method examines the most recent frame that was added to the mosaic and generates boundaries of it. The boundaries of the last frame in the overall mosaic are scaled up by three. After that, we use this portion of the mosaic in the process of feature extraction and matching. It is sufficient to guarantee that the mosaicking algorithm has the proper amount of time to execute. It now takes the same amount of time to add a new frame using this method; the time required is no longer proportional to the growing size of the mosaic.

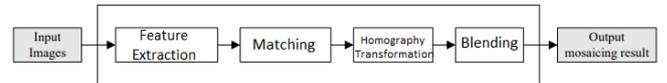

Fig. 1. Pipeline of the proposed method

### A. Feature Extraction

In order to process a digital picture, many image processing approaches require feature extraction and algorithms to find properties such as forms, edges, or movements. The stage of feature extraction in image mosaicking algorithms involves the process of extracting features from the current frame as well as the portion of the mosaic. This is done in order to match comparable key points between pictures.

As a visual feature extractor, Scale-Invariant Feature Transform (SIFT) technique is used in the proposed method. The feature points are proven to give robust matching despite being invariant to picture scale and rotation. Features extracted from image using SURF is shown in Fig. 2. Although SURF and ORB are faster [9], SIFT is preferred because SIFT can extract more and better features than these two. Additionally, because the proposed method does not use any camera parameter, there is a need for extracting better features and finding better matches to ensure mosaicking is visually pleasing. If there is a need for this process being faster, we can scale down the size of the images to accelerate the feature extraction process. In the results section we compared results of different scales of images and we compared SIFT, SURF and ORB algorithms.

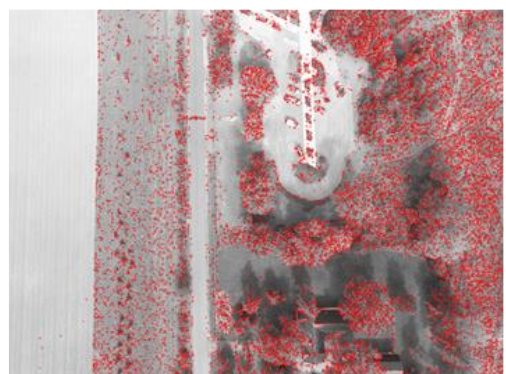

Fig. 2. Feature extraction (red points show extracted features)

### B. Matching

In the matching phase of image mosaicking algorithm, the extracted features of the current frame are often compared to those extracted from the portion of mosaic to decide appropriate location for the current frame within the mosaic.

In the proposed method, the Brute Force Matcher (BFM) algorithm is implemented to determine the association between the keypoints in the overlap region. This method does a comparison between the two sets of keypoints and matches only those whose patterns are same.

To ensure that the pairs of features provided by BFM are similar, the proposed method uses a ratio test. Essentially, the algorithm performs a distance analysis on each of the pairings provided by BFM. If the distance between each set of features falls within a manually selected ratio, the algorithm keeps the match; otherwise, it eliminates it.

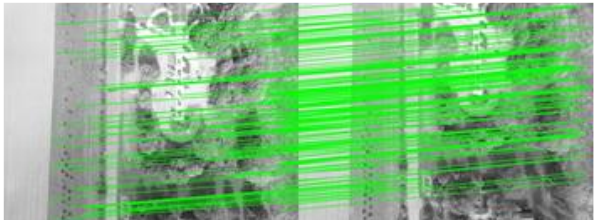

Fig. 3. Matching (Green lines matches similar features of left and right images.)

## C. Homography

Homography, is a transformation between two images of the same scene, but from a different perspective. Homography allows us to transform from one image to another image of the same scene by multiplying the homography matrix with the points in one image to find their corresponding locations in another image. In order to calculate the homography matrix, we use Direct Linear Transform algorithm [24].

RANSAC is an iterative algorithm to fit linear models. In contrast to conventional linear regressors, RANSAC is designed to be robust to outliers. In this application, the input data to RANSAC is the collection of keypoint matches that occur between current frame and the portion of the mosaic. The algorithm selects matches that are true matches (inliers) as opposed to matches that occur by accident (outliers).

Once the true matches between the two frames have been obtained, the system needs to compute the geometrical transformation by which the key points of the current frame, to be collimated with ones from the mosaic within reference system of the mosaic. Let m is the pixel with homogenous coordinates m = [ x y z ]T. Lets assume that there are two 2D images which are mosaic and current frame. It is feasible to detect the same features in them, as they are related to one another by homography H, which is a nonsingular projective matrix that maps the points from one image (m1) to the corresponding points in the other image (m2) as.

$$m_2 = Hm_1$$

$$\begin{bmatrix} x_2 \\ y_2 \\ z_2 \end{bmatrix} = \begin{bmatrix} h_{11} & h_{12} & h_{13} \\ h_{21} & h_{22} & h_{23} \\ h_{31} & h_{32} & h_{33} \end{bmatrix} \begin{bmatrix} x_1 \\ y_1 \\ z_1 \end{bmatrix}.$$

This transformation is applied to each pixel of the frame so that it can be stitched over the mosaic.

## D. Blending

After obtaining the homography transformation matrix, the algorithm applies geometric transformations to the new frame. Once the homography transformation has been applied, the new frame image is aligned with the mosaic image. After warping and aligning these images, we only need to use a technique called alpha blending to improve the transition between overlaps. In the alpha blending technique, the color in the overlapping region is determined by taking a weighted summation of the two images that overlap. The one closest to the pixel of its own frame will be assigned a greater weight. Therefore, the color transitions smoothly from one image to the next without an apparent seam.

Although there is no obvious seam, some edges in image may get hazy and those that are moving will become transparent. To counteract this blurring effect, we utilize edge detection and identify the complex regions with the most detected edges, such as areas containing trees. To make edge detection mask more useful, we apply morphological closing, which consists of a dilation followed by an erosion. If the complexity of the edges is greater, the pixel weight of the image is increased before blending is performed. Therefore, the proposed method results in mosaics that are completely smooth and exhibit no recognizable distortions. Fig.4 show a comparison of blending applied result and not applied result.

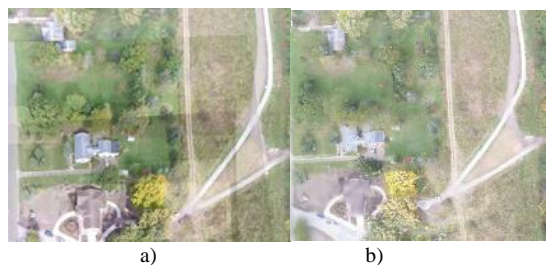

a)             b)

Fig. 4. a) unblended image b) blended image

## IV. RESULTS

In order to test the proposed image mosaicking algorithm, we used five open source datasets, which are referred to as odm-data-toledo[18], drone-dataset-sheffield-park[19], npu-phantom3-village[20], odm-data-aukerman[21], and bad-zurzach-construction [22]. These datasets include images that were taken by an unmanned aerial vehicle (UAV) flying over a farm in Toledo, Spain, a park in Florida, USA, a village in China, a park in Ohio, USA, a village in Bad Zurzach, Switzerland, respectively. odm-data-toledo has 41 images with a resolution of 4000 by 3000, drone-dataset-sheffield-park contains 77 images with a resolution of 4000 by 3000 pixels, npu-phantom3-village contains 91 images with a resolution of 1920 by 1080 pixels, odm-data-aukerman contains 32 images with a resolution of 4896 by 3672 pixels, and bad-zurzach-construction contains 37 images with a resolution of 6000 by 4000 pixels, for each image. The average flying heights were 122 m, 28 m, 165 m, 120m, and 120 m, respectively. Although the UAV was navigated with the help of GPS, we did not use the GPS data in any of proposed mosaicking method tests.

In order to compare our results, we use the tools Open Drone Map (ODM) [16] and Pix4Dmapper [17]. Open Drone Map (ODM) [16] is a toolkit that is completely open source that generates maps, point clouds, 3D models, and DEMs from drone. Pix4Dmapper is a photogrammetry software that converts images into precise and georeferenced digital models. Images captured by drone or manned aircraft are automatically converted by the software into incredibly precise georeferenced two-dimensional orthomosaics and three-dimensional models. In Fig. 5-9, results of image mosaicking for proposed method, ODM and Pix4Dmapper are shown.

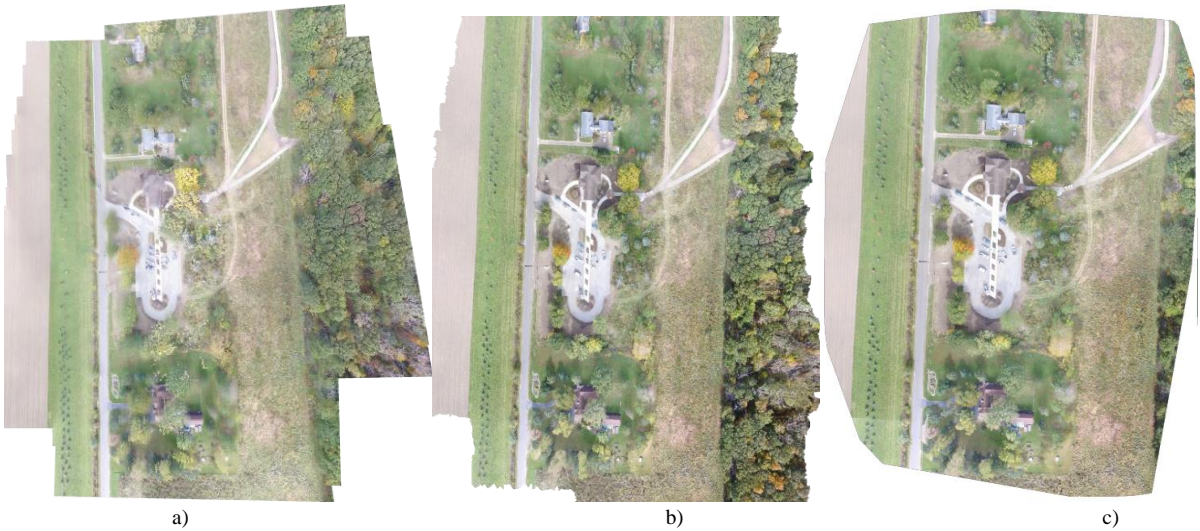

a) b) c)

Fig. 5. a) Proposed Method, b)OpenDroneMap, c) Pix4Dmapper results of odm-toledo-dataset

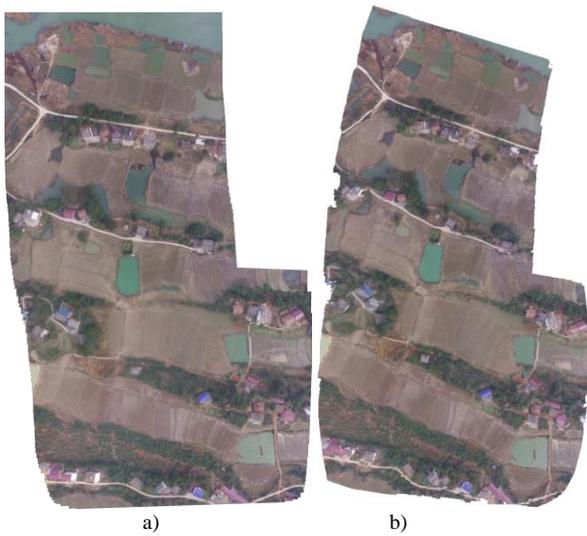

a) b)

Fig. 6. a)Proposed Method, b)Pix4Dmapper results of npu-phantom3-village

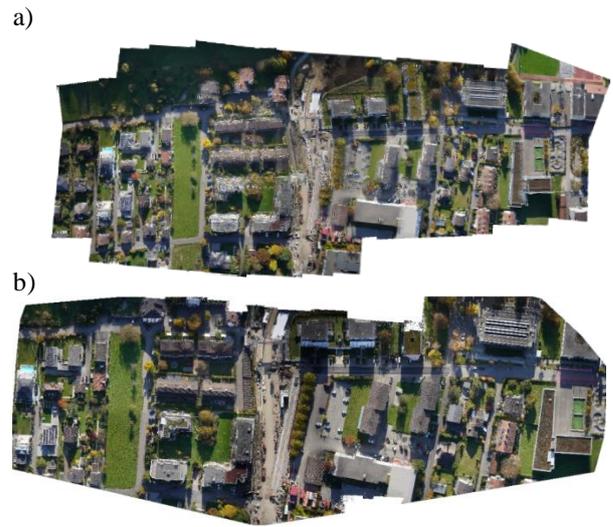

a)

b)

Fig. 7. a) Proposed Method, b) Pix4Dmapper results of bad-zurzach-construction

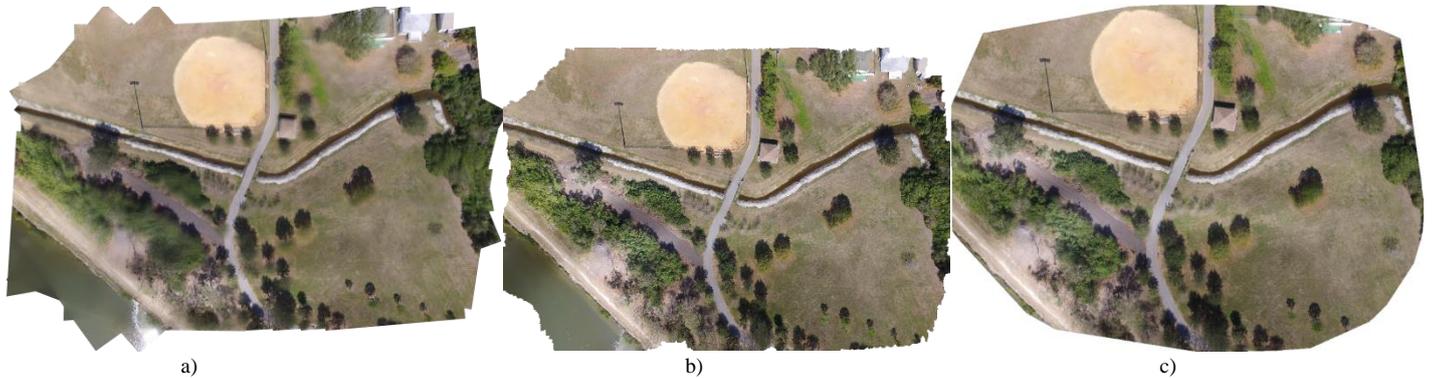

a) b) c)

Fig. 8. a) Proposed Method, b) OpenDroneMap, c) Pix4Dmapper results of drone-dataset-sheffield-park

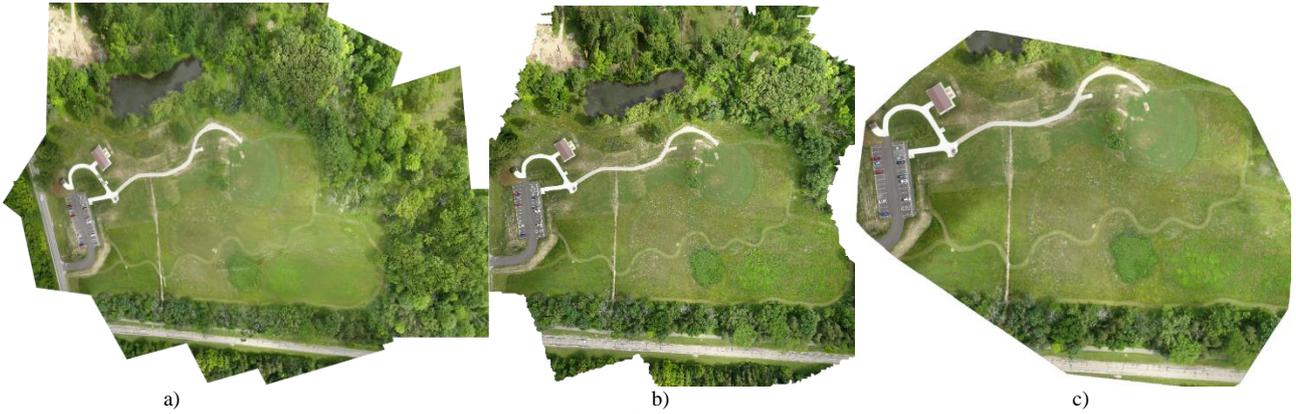

Fig.9 a) Proposed Method, b) OpenDroneMap, c) Pix4Dmapper results of odm-data-aukerman

The results demonstrate that an aesthetically pleasing result can be obtained using only the images and no additional parameters. Additionally, the results of the proposed method look better in some areas. The ODM results contain some distortions close to the house and among the trees in odm-toledo-dataset as shown in Fig. 10. On the other hand, there is not much distortion with proposed method. The proposed method can produce better results in areas with trees because it uses the blending technique that uses edge detection.

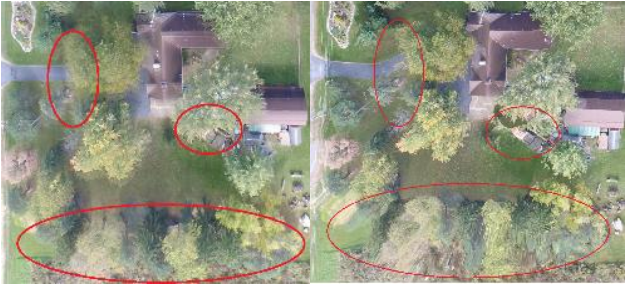

Fig. 10. Distortions (red circles) in the results by the a) proposed method b) ODM

Because ODM and Pix4Dmapper do not operate in real-time, the mosaicking process using these programs may take a considerable amount of time to finish. In contrast, the proposed method is faster than these because it is designed to process in real-time. The amount of time required to complete image mosaicking for two datasets in proposed method is displayed in Table 1.

TABLE 1. Required times to complete image mosaicking for different datasets

| Dataset | Time (s) | | |
|---|---|---|---|
| | Total | Feature Extraction | Matching |
| Odm-data-toledo | 39.89 | 10.21 | 26.24 |
| Drone-dataset-sheffield-park | 53.70 | 16.47 | 32.64 |
| Npu-phantom3-village | 47.45 | 13.32 | 29.85 |
| Odm-data-aukerman | 29.67 | 10.02 | 19.25 |
| bad-zurzach-construction | 25.47 | 9.34 | 14.72 |

The results of the feature extraction steps using SIFT, SURF and ORB are compared. As illustrated in Fig. 11, SIFT extracts more features and creates better mosaics than SURF. ORB can not extract homogeneously distributed features as SIFT, but it can do so 3 times as fast (Table 2). Because it does not extract homogeneously distributed features, mosaicking using ORB becomes distorted after 4th image is stitched (Fig. 12). Sometimes, a fast algorithm is required for real-time processing. If it is necessary, we can reduce the size of images to speed up the process.

TABLE 2. Required times of the first stiching of mosaicking for different feature extraction alghorithms.

| Algorithm | Number of Features Extracted | Time of the first stitching (s) | | |
|---|---|---|---|---|
| | | Total | Feature Extraction | Matching |
| SIFT | 6704 | 4.5024 | 1.0220 | 3.4186 |
| ORB | 9973 | 1.3454 | 0.6501 | 0.6838 |
| SURF | 3927 | 1.8963 | 0.8853 | 0.9714 |

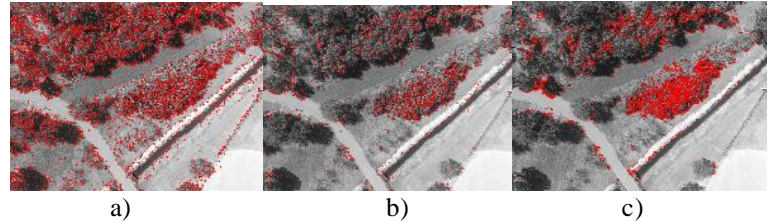

Fig. 11. Extracted features using a) SIFT, b) SURF, c) ORB

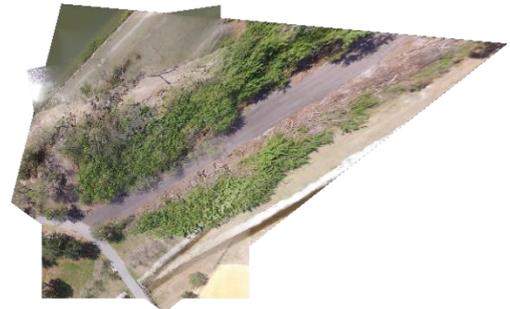

Fig.12 Distorted image mosaicking using ORB

When image sizes are large, the feature extraction stage can sometimes take longer. Images may require to be scaled down in these circumstances. Feature extraction by using SIFT algorithm is invariant to image scale. Scaling so does not significantly interfere with the process of creating an image mosaic. We compared the results of images at different scales to make sure that scaling is not particularly crucial for creating pleasing image mosaics. As seen in Table 3, if we scale down the images, we can create mosaics faster. However, when size of images are reduced, we may forsake highly detailed images. We can use small-scaled images if the goal of the image mosaicking does not require detailed pixels. Table 3 shows the required times for mosaicking of different size of images in toledo-dataset.

TABLE 3. Required times of the first stiching of image mosaicking for different size of images

| Scale | Time (s) | | | |
|---|---|---|---|---|
| | *Total* | *Feature Extraction* | *Matching* | *Stitching* |
| 100% | 4.50 | 1.02 | 2.89 | 0.59 |
| 50% | 0.71 | 0.39 | 0.15 | 0.17 |
| 25% | 0.18 | 0.12 | 0.01 | 0.05 |

The mosaicking results produced by the proposed method are found to be aesthetically pleasing for each of the datasets. The proposed method yields seamless mosaics that do not exhibit obvious distortion in any of the result. Experiments show that when there is no demand for high accuracy mosaicking, especially when GPS data are either unavailable or not aligned to the imagery, the approach presented provides an effective and practical alternative method for real-time mosaicking of the UAV imagery.

For featureless scenes (e.g., flat areas, surface of river and lakes), the image matching algorithm may fail since the proposed method is image-based. Future research will concentrate on the issues regarding the improvement of mosaicking, such as how to improve blending process more to get rid of blur effects entirely. In addition, we will investigate how to adapt the method to mountainous regions.

As future work, we plan to extend this work to images from multispectral sensors containing rural areas. This will allow multispectral analysis of agriculture and infrastructure to be performed on a larger scale allowing efficient consolidation of useful information. For this task, further optimization of feature representations and optimization of multispectral inputs and outputs for homography calculation and blending operations.